# Cross-Model Consistency of AI-Generated Exercise Prescriptions: A Repeated Generation Study Across Three Large Language Models

Kihyuk Lee*

Data Convergence Team, Office of Hospital Information, Seoul National University Bundang Hospital, Seongnam, Republic of Korea

**Correspondence to**
Kihyuk Lee, PhD
Data Convergence Team, Office of Hospital Information, Seoul National University Bundang Hospital, 172 Dolma-ro, Bundang-gu, Seongnam, 13605, Republic of Korea

**Highlights**

- Temperature=0 conditions produced markedly different output profiles across GPT-4.1, Claude Sonnet 4.6, and Gemini 2.5 Flash
- GPT-4.1 achieved both textual diversity and semantic consistency simultaneously across all six clinical scenarios
- Gemini 2.5 Flash showed pronounced output repetition, with only 27.5% unique outputs across 120 generations
- Safety expression scores reached ceiling levels across all three models, limiting their utility as a differentiating evaluation metric
- Model selection directly affects exercise prescription consistency and should be treated as a clinical rather than technical decision

**Abstract**

This study compared repeated generation consistency of exercise prescription outputs across three large language models (LLMs), specifically GPT-4.1, Claude Sonnet 4.6, and Gemini 2.5 Flash, under temperature=0 conditions. Each model generated prescriptions for six clinical scenarios 20 times, yielding 360 total outputs analyzed across four dimensions: semantic similarity, output reproducibility, FITT classification, and safety expression. Mean semantic similarity was highest for GPT-4.1 (0.955), followed by Gemini 2.5 Flash (0.950) and Claude Sonnet 4.6 (0.903), with significant inter-model differences confirmed ($H = 458.41$, $p < .001$). Critically, these scores reflected fundamentally different generative behaviors: GPT-4.1 produced entirely unique outputs (100%) with stable semantic content, while Gemini 2.5 Flash showed pronounced output repetition (27.5% unique outputs), indicating that its high similarity score derived from text duplication rather than consistent reasoning. Identical decoding settings thus yielded fundamentally different consistency profiles, a distinction that single-output evaluations cannot capture. Safety expression reached ceiling levels across all models, confirming its limited utility as a differentiating metric. These results indicate that model selection constitutes a clinical rather than merely technical decision, and that output behavior under repeated generation conditions should be treated as a core criterion for reliable deployment of LLM-based exercise prescription systems.

## 1. Introduction

The rapid advancement of large language models (LLMs) has opened new possibilities in healthcare and health management (Raza et al., 2024; Meng et al., 2024). LLMs can generate contextually appropriate text based on user prompts, and their potential applications have been discussed across a range of domains including clinical consultation support, health information provision, behavior change facilitation, and personalized exercise advice (Aydin et al., 2024; Zhang & Liu, 2024; Philuek et al., 2025; Negra et al., 2026). In the fields of exercise science and sports medicine in particular, LLMs have attracted attention as tools for supporting exercise guidance for non-specialist users and assisting in the development of individualized exercise programs (Li et al., 2025; Enichen et al., 2025).

Exercise prescription is, however, fundamentally distinct from the general provision of exercise-related information (Garber et al., 2011; Festa et al., 2023). It constitutes a specialized decision-making process that requires comprehensive consideration of an individual's health status, disease characteristics, functional capacity, medications, exercise contraindications, and precautionary criteria. A growing body of research has sought to evaluate the practical value of AI-generated exercise prescriptions in this domain. LLM outputs have been assessed by expert evaluation across a range of contexts, including resistance training programs, running training plans, and prescriptions for patients with type 2 diabetes (Washif et al., 2024; Düking et al., 2024; Akrimi et al., 2025). In the area of cardiovascular health, multiple LLMs demonstrated generally high levels of guideline adherence, although performance differences were observed across models (Nduka et al., 2025). Progress has also been made in the development of AI-assisted evaluation tools; an Adaptive Precision Boolean Rubric (Adaptive-PBR) based on ACSM guidelines demonstrated superior inter-rater reliability compared to conventional Likert scales (ICC = 0.83) and a 63% reduction in evaluation time (Lai et al., 2026). A recent scoping review confirmed that this body of research is transitioning from an exploratory to a more systematic validation phase (Lai et al., 2025a).

Within this context, the present research team has conducted a series of studies proposing output reliability as a novel evaluation dimension for LLM-based exercise prescription. Choi et al. (2026) applied a structured three-stage prompt framework to three clinical cases involving complex clinical considerations, analyzing Gemini 2.5-generated prescriptions using an expert evaluation rubric. While prompt structuring was associated with improvements in safety and guideline adherence scores, inter-rater reliability was low (ICC(2,3) = 0.139), and the study identified the difficulty of ensuring complete statistical independence across repeated outputs under identical conditions as a limitation, proposing the systematic analysis of output consistency as a direction for future research. Building on this, a subsequent study extended the same prompt framework to six clinical cases with 20 repeated generations per case, quantifying intra-model consistency along three dimensions: semantic similarity (SBERT cosine similarity), FITT component classification, and safety expression inclusion. Results showed that semantic similarity was maintained at high levels across all cases (mean cosine similarity: 0.879–0.939), yet meaningful variability was observed in key quantitative components such as exercise intensity, and although safety-related expressions were present in all outputs, their frequency differed significantly across cases ($H = 86.18$, $p < 0.001$). These studies are notable for introducing reproducibility and consistency as independent evaluation dimensions beyond the conventional accuracy- and safety-centered paradigm; however, whether the observed variability reflects characteristics unique to a single model or properties common across LLMs more broadly remains an unresolved question.

LLMs are inherently variable due to their probabilistic token generation mechanism, and meaningful differences in output can occur across repeated generations from identical prompts (Shyr et al., 2025). Previous research has noted that evaluation strategies in the field of LLM-based exercise and health coaching are generally fragmented and lack a standardized framework (Lai et al., 2025b), highlighting the need to establish a methodological basis for comparing output consistency across models. Whether such variability reflects characteristics unique to specific model architectures or represents a general property of LLMs remains unclear. Despite growing evidence in clinical AI research that model selection influences output quality and reliability (Croxford et al., 2025), systematic cross-model comparisons of exercise prescription output consistency have not been sufficiently conducted to date.

The present study aims to address this gap by systematically comparing intra- and inter-model consistency across three widely used LLMs, specifically GPT-4.1 (OpenAI), Claude Sonnet 4.6 (Anthropic), and Gemini 2.5 Flash (Google), applying the same clinical cases and prompt protocol established in previous studies.

## 2. Materials and Methods

### *2.1. Study Design*

This study employed an experimental observational design to compare repeated generation consistency of exercise prescription outputs across three LLMs. Identical clinical scenarios and prompts were submitted to each model under controlled conditions, and intra-model consistency and inter-model differences were quantitatively evaluated. The design extends a previous single-model study (Choi et al., 2026) to directly compare the effect of model selection on exercise prescription consistency. The overall study design is illustrated in Figure 1, which summarizes the four-phase workflow: clinical scenario input, repeated generation across three LLMs, AI-as-a-Judge evaluation using Claude Sonnet 4.6, and multi-dimensional consistency analysis.

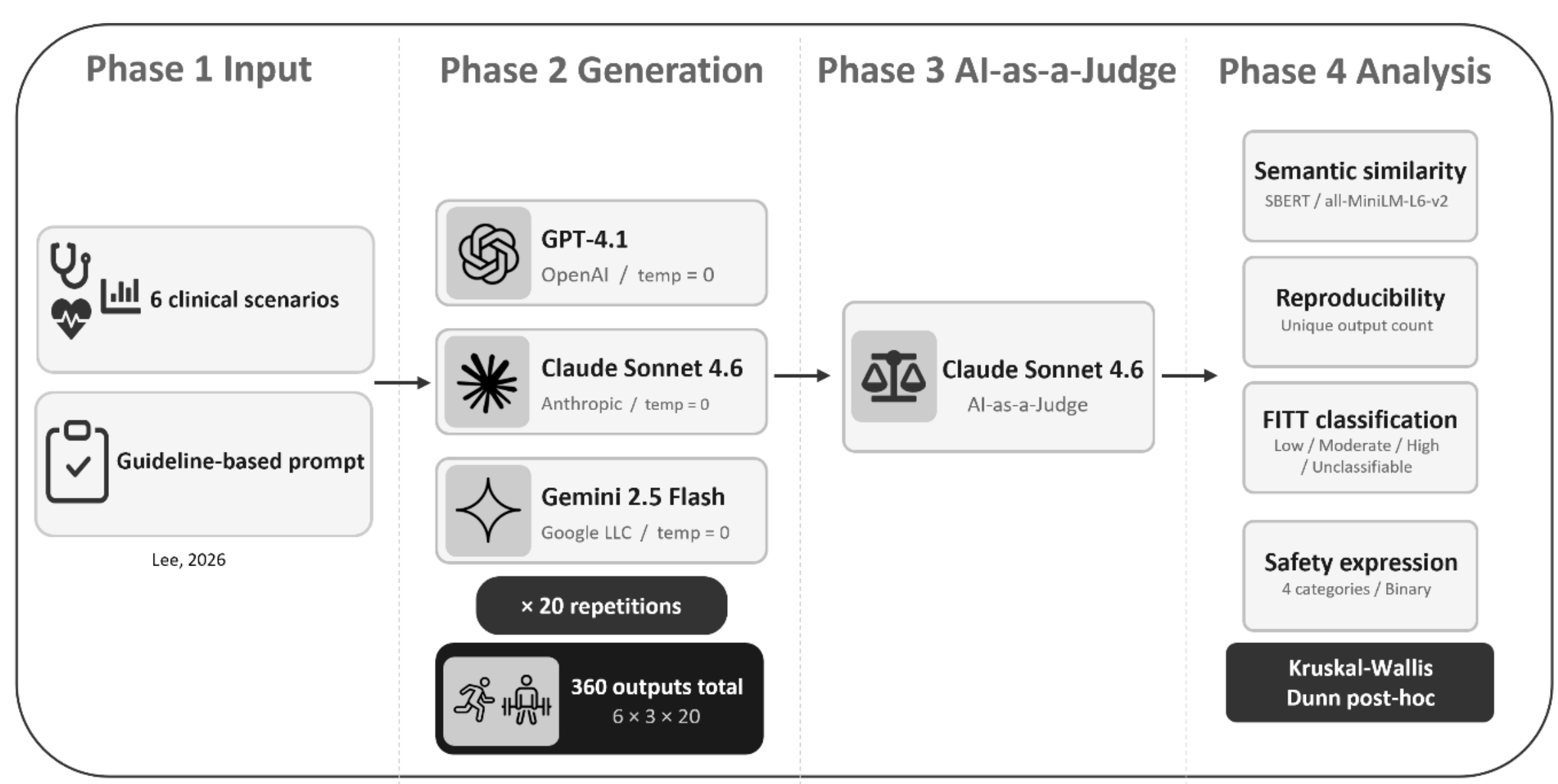

**Figure 1.** Study design overview.

### *2.2. Clinical Scenarios*

Six clinical scenarios were used, comprising three cases from a previous study (Choi et al., 2026) and three additional cases introduced in Lee et al. (2026), covering a range of prescription contexts from high-risk clinical presentations to healthy adult cases. Scenario details are identical to those described in Lee et al. (2026).

### *2.3. Model Selection and Experimental Conditions*

GPT-4.1 (OpenAI), Claude Sonnet 4.6 (Anthropic), and Gemini 2.5 Flash (Google LLC) were selected as generator models, representing major commercially available LLMs from independent model families. API versions and conditions are presented in Table 1. Temperature was fixed at 0 across all models to suppress stochastic variability, with all other parameters kept at API defaults. Outputs were collected via automated API calls, ensuring session equivalence. No chain-of-thought or other reasoning elicitation techniques were applied.

### *2.4. Prompt and Output Generation*

A guideline-based prompt developed and validated in previous studies (Lee et al., 2026; Choi et al., 2026) was applied identically across all three models, ensuring that output differences reflect model-specific generative characteristics rather than input variation. The full prompt is provided in Supplementary Material 1. Each scenario was generated 20 times per model, yielding 360 total outputs (6 scenarios × 20 repetitions × 3 models), following repeated generation designs for estimating LLM output variability (Shyr et al., 2025).

### *2.5. Output Preprocessing*

Outputs were collected in their original form and preprocessed to remove non-prescription elements such as follow-up prompts, emoticons, and decorative symbols. A standardized extraction prompt using Claude Sonnet 4.6 was applied uniformly across all 360 outputs to extract prescription content without altering semantic content. Preprocessing accuracy was verified through manual review of 36 randomly sampled outputs (10%) by the first author.

### *2.6. Consistency Evaluation*

Consistency was evaluated along three dimensions: semantic consistency, structural consistency, and safety consistency, each calculated at the scenario and model level.

Semantic consistency was assessed using pairwise cosine similarity across all 190 pairs of the 20 outputs per scenario-model condition, computed with a pretrained sentence embedding model (all-MiniLM-L6-v2; Reimers & Gurevych, 2019). Mean and standard deviation of the similarity distribution served as consistency indices.

Structural consistency was evaluated based on the four FITT components using Claude Sonnet 4.6 as a separate evaluator model to prevent self-evaluation bias (Zheng et al., 2023; Li et al., 2024). Intensity was classified as low, moderate, or high for initial prescription weeks 1 to 4 following the criteria described in Lee et al. (2026), which were based on ACSM guidelines (Garber et al., 2011; American College of Sports Medicine, 2024), an expert consensus statement (Bishop et al., 2025), and the repetition continuum framework (Schoenfeld et al., 2021; Currier et al., 2026). Outputs with insufficient information were labeled "estimated" or "unclassifiable" and reported separately.

Safety consistency was evaluated by binary judgment of four categories, specifically contraindications, precautions, monitoring, and warnings, using Claude Sonnet 4.6 as evaluator, with category definitions based on Choi et al. (2026) and ACSM guidelines (American College of Sports Medicine, 2024). Inclusion rates (%) were calculated by scenario and model. The use of Claude Sonnet 4.6 as both generator and evaluator is acknowledged as a potential source of self-evaluation bias; however, as the task was limited to binary judgment with predefined criteria, which likely reduces subjective interpretation.

The full Claude Judge prompts and classification criteria are provided in Supplementary Material 2. Output reproducibility was evaluated by exact match comparison of preprocessed texts within each scenario-model condition, with unique output counts and proportions (%) calculated by scenario and model.

*2.7. Statistical Analysis*

Descriptive statistics are presented as mean ± standard deviation. Non-parametric tests were applied throughout given the small sample size per scenario ($n = 20$). For RQ1, semantic consistency distributions, FITT classifications, and safety inclusion rates were summarized descriptively, and cross-scenario differences were tested using the Kruskal-Wallis test. For RQ2, cross-model differences in the same measures were compared using the Kruskal-Wallis test, followed by pairwise comparisons using the Mann-Whitney U test. Significant results in both analyses were followed by Dunn post-hoc tests with Bonferroni correction. The significance level was set at $p < 0.05$. FITT and safety outcomes were additionally summarized as frequencies and proportions and compared descriptively across models. All analyses were conducted in Python 3.12.13 (Google Colaboratory) using the sentence-transformers, scipy, and scikit_posthocs libraries.

## 3. Results

*3.1 Intra-model Semantic Consistency (RQ1)*

Overall mean semantic similarity was highest for GPT-4.1 (Mean = 0.955, SD = 0.028), followed by Gemini-2.5-Flash (Mean = 0.950, SD = 0.070) and Claude-Sonnet-4.6 (Mean = 0.903, SD = 0.071). All three models showed significant variation in consistency across scenarios (all $p < 0.001$), with detailed pairwise comparisons presented in Table 1 and Figure 2. Across all three models, the highest consistency was observed in S2 (knee osteoarthritis with fall risk), suggesting a tendency for outputs to converge when clinical constraints are clearly defined. Notably, in S6, Claude-Sonnet-4.6 exceeded

GPT-4.1 in mean similarity (0.924 vs. 0.911), indicating that model consistency rankings are not uniform across scenarios and may vary depending on clinical context.

**Table 1.** Intra-model Semantic Consistency: SBERT Cosine Similarity by Model and Scenario

| Scenario | GPT-4.1 | Claude-Sonnet-4.6 | Gemini-2.5-Flash |
|---|---|---|---|
| S1 | 0.949 ± 0.033 | 0.878 ± 0.074 | 0.951 ± 0.078 |
| S2 | 0.984 ± 0.010 | 0.952 ± 0.033 | 0.984 ± 0.048 |
| S3 | 0.960 ± 0.024 | 0.896 ± 0.084 | 0.925 ± 0.076 |
| S4 | 0.967 ± 0.023 | 0.878 ± 0.110 | 0.946 ± 0.074 |
| S5 | 0.960 ± 0.027 | 0.890 ± 0.079 | 0.913 ± 0.109 |
| S6 | 0.911 ± 0.049 | 0.924 ± 0.047 | 0.979 ± 0.035 |
| Total | 0.955 ± 0.028 | 0.903 ± 0.071 | 0.950 ± 0.070 |
| K-W H | 286.02 | 104.37 | 125.05 |
| K-W p | < 0.001 | < 0.001 | < 0.001 |

Mean ± SD based on 190 pairwise comparisons per scenario (20 trials); K-W: Kruskal-Wallis.

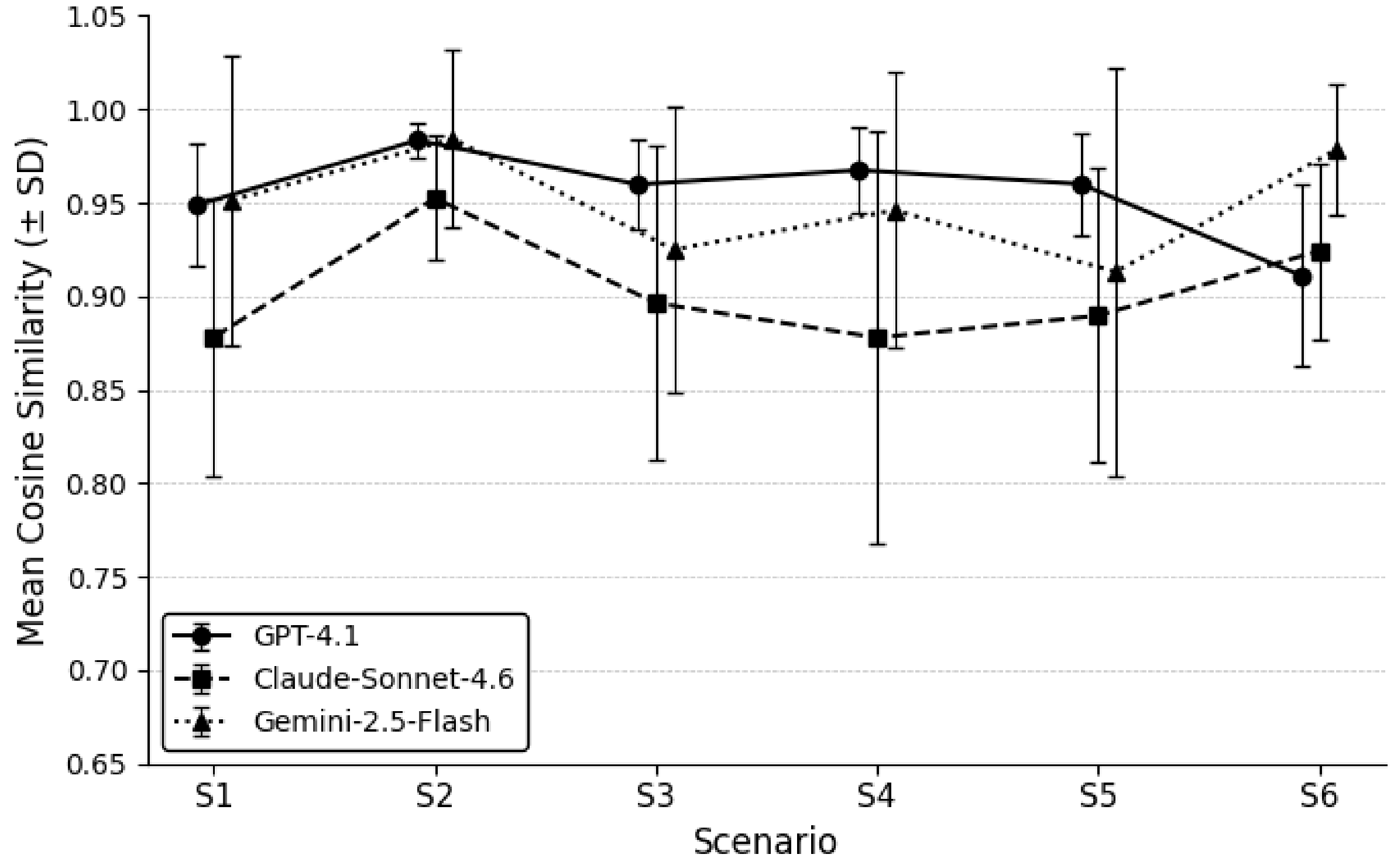

**Figure 2.** Intra-model Semantic Consistency Across Scenarios by Model

Mean cosine similarity (± SD) for each model across six clinical scenarios (S1–S6), based on 190 pairwise comparisons per scenario (20 trials). Higher values indicate greater output consistency within each model. All three models showed significant variation across scenarios (Kruskal-Wallis, all $p < 0.001$).

*3.2 Inter-model Semantic Consistency (RQ2)*

A Kruskal-Wallis test confirmed significant overall differences across the three models ($H = 458.41$, $p < 0.001$). Mann-Whitney U post hoc tests with Bonferroni correction ($\alpha = 0.0167$) revealed significant differences between all three model pairs: GPT-4.1 vs. Claude-Sonnet-4.6 ($U = 873{,}166$, $p < 0.001$), GPT-4.1 vs. Gemini-2.5-Flash ($U = 450{,}891$, $p < 0.001$), and Claude-Sonnet-4.6 vs. Gemini-2.5-Flash ($U = 357{,}035$, $p < 0.001$). Scenario-level comparisons also revealed significant differences across all six scenarios ($H = 33.06$–$216.28$, all $p < 0.001$). Dunn's post hoc test showed that Claude-Sonnet-4.6 was significantly distinguished from the other two models in all scenarios, while GPT-4.1 and Gemini-2.5-Flash showed no significant difference in S3 ($p = 0.197$) and S5 ($p = 0.425$). Detailed results are presented in Tables 2 and 3.

**Table 2.** Overall Inter-model Comparison of SBERT Cosine Similarity

| Comparison | U | p |
|---|---|---|
| GPT-4.1 vs. Claude-Sonnet-4.6 | 873,166 | < 0.001 |
| GPT-4.1 vs. Gemini-2.5-Flash | 450,891 | < 0.001 |
| Claude-Sonnet-4.6 vs. Gemini-2.5-Flash | 357,035 | < 0.001 |

Comparisons based on 1,140 pairwise similarity values per model (190 pairs × 6 scenarios).

**Table 3.** Scenario-level Inter-model Comparison of SBERT Cosine Similarity

| Scenario | K-W H | Dunn's post hoc p-value (Bonferroni-corrected) | | |
|---|---|---|---|---|
| | | GPT-4.1 vs. Claude-Sonnet-4.6 | GPT-4.1 vs. Gemini-2.5-Flash | Claude-Sonnet-4.6 vs. Gemini-2.5-Flash |
| S1 | 157.07 | < 0.001 | < 0.001 | < 0.001 |
| S2 | 172.05 | 0.004 | < 0.001 | < 0.001 |
| S3 | 33.06 | < 0.001 | 0.197 | < 0.001 |
| S4 | 88.44 | < 0.001 | 0.001 | < 0.001 |
| S5 | 59.09 | < 0.001 | 0.425 | < 0.001 |
| S6 | 216.28 | 0.018 | < 0.001 | < 0.001 |

K-W: Kruskal-Wallis. Kruskal-Wallis test was significant in all scenarios (all $p < 0.001$). Dunn's post hoc p-values

are Bonferroni-corrected (α = 0.0167).

*3.3 Output Reproducibility*

The proportion of unique outputs differed markedly across models under temperature = 0 conditions: GPT-4.1 produced entirely unique outputs across all 120 generations (100.0%), followed by Claude-Sonnet-4.6 (84/120, 70.0%) and Gemini-2.5-Flash (33/120, 27.5%) (Table 4). Gemini-2.5-Flash showed the strongest duplication pattern, with S2 producing only 2 unique outputs across 20 repetitions.

**Table 4.** Unique Output Counts by Model and Scenario

| Scenario | GPT-4.1 (n=20) | Claude-Sonnet-4.6 (n=20) | Gemini-2.5-Flash (n=20) |
|---|---|---|---|
| S1 | 20 | 20 | 5 |
| S2 | 20 | 12 | 2 |
| S3 | 20 | 14 | 8 |
| S4 | 20 | 14 | 5 |
| S5 | 20 | 13 | 9 |
| S6 | 20 | 11 | 4 |
| Total | 120 (100.0%) | 84 (70.0%) | 33 (27.5%) |

S1–S6: Clinical scenarios. Each scenario was repeated 20 times per model (total 120 trials per model). Values indicate the number of unique outputs observed.

*3.4 FITT Classification*

FITT classification results are summarized in Table 5. Low-intensity aerobic exercise was dominant across clinical cases in GPT-4.1 and Claude-Sonnet-4.6 (83.8% and 96.2%, respectively), while Gemini-2.5-Flash showed a notably higher unclassifiable rate in the clinical group (Aerobic: 26.2%, Resistance: 32.5%). Moderate- and high-intensity prescriptions increased in healthy adult cases across all three models. A notable exception was observed in Gemini-2.5-Flash, which produced unclassifiable resistance exercise intensity outputs at a substantially higher rate in clinical scenarios (32.5%), with the majority concentrated in S4 (n = 17/20, 85.0%).

**Table 5.** Exercise Intensity Classification by Model and Clinical Group (Initial Phase, Weeks 1–4)

| Component | Group | GPT-4.1 | Claude-Sonnet-4.6 | Gemini-2.5-Flash |
|---|---|---|---|---|
| Frequency | S1–S4 (Clinical) | 3 days/week (57.5%) | 3 days/week (41.2%) | 3–4 days/week |

| Component | Group | GPT-4.1 | Claude-Sonnet-4.6 | Gemini-2.5-Flash |
|---|---|---|---|---|
| | S5–S6 (Healthy) | 4 days/week (47.5%) | 6 days/week (50.0%) | 4–5 days/week |
| Aerobic Intensity | S1–S4 (Clinical) | Low 83.8%, Moderate 15.0%, Unc 1.2% | Low 96.2%, Moderate 1.2%, Unc 2.5% | Low 67.5%, Moderate 6.2%, Unc 26.2% |
| | S5–S6 (Healthy) | Moderate 95.0%, High 2.5%, Unc 2.5% | Moderate 92.5%, High 2.5%, Unc 5.0% | Moderate 100% |
| Resistance Intensity | S1–S4 (Clinical) | Low 98.8%, Moderate 1.2% | Low 97.5%, Moderate 2.5% | Low 67.5%, Unc 32.5% |
| | S5–S6 (Healthy) | Low 2.5%, Moderate 92.5%, High 2.5%, Unc 2.5% | Low 15.0%, Moderate 75.0%, High 10.0% | Moderate 95.0%, High 5.0% |
| Time | S1–S4 (Clinical) | 15–20 min | 20–30 min | 10–20 min |
| | S5–S6 (Healthy) | 30–75 min | 35–75 min | 30–75 min |

Frequency and Time: most frequently observed pattern. Unc: unclassifiable. Values represent proportions of 80 trials (S1–S4) and 40 trials (S5–S6) per model.

### *3.5 Safety Expression*

All three models showed ceiling-level safety expression scores, with mean Safety_Total scores of 3.99 (SD = 0.09), 3.97 (SD = 0.22), and 3.93 (SD = 0.25) for Gemini-2.5-Flash, Claude-Sonnet-4.6, and GPT-4.1, respectively. Contraindications were present in 100% of outputs across all models, and inclusion rates for precautions, monitoring, and warnings were similarly high (Table 6). Given the ceiling-level scores across all models, formal statistical comparison was not conducted.

**Table 6.** Safety Expression Inclusion Rate by Model (Binary Classification)

| Model | Inclusion Rate, n (%) | | | | Safety Total Mean ± SD |
|---|---|---|---|---|---|
| | Contraindication | Precaution | Monitoring | Warning | |
| GPT-4.1 | 120 (100.0) | 120 (100.0) | 119 (99.2) | 113 (94.2) | 3.93 ± 0.25 |
| Claude-Sonnet-4.6 | 120 (100.0) | 119 (99.2) | 120 (100.0) | 117 (97.5) | 3.97 ± 0.22 |
| Gemini-2.5-Flash | 120 (100.0) | 120 (100.0) | 120 (100.0) | 119 (99.2) | 3.99 ± 0.09 |

n: number of outputs including the safety category out of 120 total outputs per model. Safety Total: sum of four binary items (range 0–4).

## 4. Discussion

Previous studies on LLM-based exercise prescription have largely focused on single-model evaluations, and cross-model differences in output characteristics have not been adequately examined. Although a previous study confirmed the intra-model consistency of Gemini under repeated generation conditions (Lee et al., 2026), whether similar patterns emerge across other models remains unclear. The present study addresses this gap by evaluating not only output accuracy but also output consistency under repeated generation conditions and its underlying characteristics. In particular, it seeks to distinguish whether observed consistency reflects stable reasoning or derives from superficial output repetition. This distinction carries direct implications for clinical deployment.

All three models showed significant variation in semantic similarity across scenarios (all $p < .001$), yet mean similarity levels differed by model. GPT-4.1 produced the highest mean similarity (0.955), followed by Gemini 2.5 Flash (0.950) and Claude Sonnet 4.6 (0.903). The fact that such divergent output profiles emerged under identical prompt and temperature conditions confirms that prescription consistency is a model-dependent property, not merely a function of decoding parameters. Previous studies reported that GPT-4 outperformed Gemini in exercise prescription quality (Puce et al., 2025; Nduka et al., 2025); however, those findings were based on expert evaluations of single outputs. The present results extend this evidence by demonstrating that cross-model differences persist not only in output quality but in output behavior under repeated generation, a dimension that single-output evaluations cannot capture.

GPT-4.1 and Gemini 2.5 Flash did not show significant differences in semantic similarity in Scenario 3 (cancer patient) and Scenario 5 (healthy adult female) ($p = 0.197$, $p = 0.425$). The convergence of inter-model differences under highly constrained or simple prescription demands suggests that LLM output variability is linked to clinical complexity, consistent with the limitations of AI-based exercise prescription in complex clinical contexts noted by Kim et al. (2026). Claude, by contrast, was significantly distinguished from the other two models across all scenarios, suggesting a systematic difference in generative behavior rather than scenario-specific variation.

Output reproducibility differed markedly across the three models under identical temperature=0 conditions. GPT-4.1 produced entirely unique outputs across all 120 generations (100%), while Claude Sonnet 4.6 produced 84 unique outputs (70.0%) and Gemini 2.5 Flash only 33 (27.5%). These results empirically confirm that temperature=0 does not guarantee output determinism in clinical exercise prescription contexts, consistent with previous observations that infrastructure-level variables such as floating-point non-associativity and batch size variation can prevent complete reproducibility even at zero temperature (Atil et al., 2025; Song et al., 2024).

The pattern observed in Gemini 2.5 Flash warrants particular attention. In Scenario 2 (knee osteoarthritis with fall risk), only 2 unique outputs were produced across 20 repetitions, a pattern confirmed upon replication (S2: 11 unique; S6: 5 unique). Although this degree of repetition produces a high aggregate similarity score (0.950), the consistency reflected here derives from output repetition rather than stable semantic reasoning, a distinction that aggregate similarity metrics alone cannot reveal. Previous studies have noted that LLM-generated exercise prescriptions tend to lack specificity and individualization (Washif et al., 2024; Dergaa et al., 2024); the present findings suggest that this problem may manifest in a more structural form in certain models. If prescription outputs remain effectively fixed regardless of variation in patient input, this may represent a structural limitation for personalization that cannot be resolved through prompt engineering or parameter adjustment alone.

GPT-4.1 showed a contrasting profile. Its outputs varied in surface expression across repetitions, yet semantic similarity was the highest among the three models (0.955). This pattern, varying in form while maintaining stable core content, represents the output characteristic most compatible with reliable clinical deployment: avoiding both the user distrust associated with verbatim repetition and the safety concerns raised by semantic instability.

FITT classification results were broadly consistent with ACSM guidelines, with low-intensity exercise dominating clinical scenarios and moderate- to high-intensity prescriptions increasing in healthy adult cases (ACSM, 2024). A tendency for LLMs to produce safety-oriented, conservative outputs for clinical populations has been reported in previous studies (Kim, 2026), and at the level of intensity categorization, all three models appeared to maintain clinically appropriate judgment. However, resistance exercise intensity outputs from Gemini 2.5 Flash were unclassifiable in 32.5% of clinical trials overall, with the pattern concentrated in Scenario 4 (n = 17/20, 85%). Resistance exercise intensity must be specified in quantitative terms, such as %1RM or RPE, to be clinically applicable; its systematic absence indicates that outputs may lack critical clinical content even when they appear structurally complete. A comparable pattern of maintained safety with insufficient comprehensiveness has been reported in LLM-based exercise prescription for rare neuromuscular conditions (Schütze et al., 2026).

All three models showed ceiling-level safety expression scores ranging from 3.93 to 3.99 out of 4.00, with no formal statistical comparison conducted given the ceiling-level distribution. This suggests that current-generation LLMs consistently incorporate safety-related language in exercise prescription outputs regardless of model architecture, and that safety expression alone is insufficient to differentiate models in terms of clinical utility. This finding is consistent with the broader literature. Zaleski et al. (2024) reported that ChatGPT generally included safety-related content across 26 clinical populations but lacked comprehensiveness, and Nduka et al. (2025) similarly found that all four models generated appropriate safety recommendations. Taken together, these findings suggest that safety expression has already converged to a baseline level across current-generation LLMs, limiting its utility as a differentiating evaluation metric.

The ceiling effect simultaneously reveals a methodological limitation of safety expression scoring. Capturing the presence of cautionary language is a fundamentally different task from evaluating whether that language is contextually appropriate and integrated with the prescription content. A concrete illustration is provided by Scenario 4: despite 85% of Gemini's resistance intensity outputs being unclassifiable, safety expression scores remained at ceiling, demonstrating that the presence of safety language does not necessarily correspond to clinical completeness of the prescription.

Future evaluation frameworks should incorporate more granular criteria beyond the presence of safety language, including scenario-specific contraindication accuracy and content-prescription alignment. Previous studies employing AI-assisted adaptive rubrics have similarly noted the limitations of Likert-based safety evaluation (Lai et al., 2026), and the need for more objective and differentiated assessment systems has emerged as a shared challenge in this field. This issue warrants attention in the standardized evaluation framework currently under development as part of the present research series.

The present findings have direct implications for clinicians and developers considering the integration of LLMs into exercise prescription workflows. As illustrated most clearly by the contrast between GPT-4.1 and Gemini-2.5-Flash, output reproducibility and semantic consistency are distinct constructs, and temperature settings alone cannot guarantee deployment reliability. Clinicians should therefore treat decoding parameters not as guarantees of consistency but as one factor among many, and empirically evaluate model-specific output characteristics under real-world use conditions. The tendency for output repetition to intensify in complex clinical cases further suggests that greater caution is warranted in model selection and output verification for patients with higher clinical complexity.

More broadly, this study contributes to building a methodological foundation for the systematic evaluation of LLMs in sports and exercise science. As efforts to integrate LLM-based exercise prescription into conversational interfaces and personalized platforms continue to grow (Shin et al., 2025; He et al., 2026), the question of which model produces more consistent prescriptions, rather than simply better ones, should be recognized as a core criterion of clinical reliability. Together with a previous study examining intra-model consistency under probabilistic conditions (Lee et al., 2026) and an earlier study validating the safety and guideline adherence of LLM-generated exercise prescriptions (Choi et al., 2026), the present study is part of an ongoing research effort to accumulate evidence toward a standardized evaluation framework for AI-generated exercise prescription.

The present study has several limitations. First, all analyses were conducted under a single condition of temperature=0. In real-world clinical practice, LLM-based exercise prescription may occur across a range of temperature settings, user-defined prompts, and conversational contexts, and such variations may meaningfully affect output characteristics. Second, the study evaluated specific model versions at a single point in time; given the rapid pace of LLM development, the findings may not generalize to subsequent model versions. Third, while SBERT-based semantic similarity captures meaning at the level of sentence embeddings, outcomes such as clinical accuracy, guideline adherence, and patient safety can only be directly verified through expert clinical evaluation. Fourth, the six clinical scenarios used in this study do not represent the full complexity of exercise prescription contexts encountered in sports medicine and clinical practice. Fifth, the prompt structure was held constant throughout, making it difficult to assess how differences in prompt design might influence cross-model variation. Prompt specificity has been shown to significantly affect LLM output quality (Puce et al., 2025), and future research should include cross-model comparisons under varied prompting conditions. Sixth, the inter-model differences observed in this study may be partly dependent on the specific prompt and scenario configurations employed, and different results may emerge under alternative prompt designs or clinical contexts.

## 5. Conclusion

The present study confirmed that LLM-generated exercise prescription outputs differ markedly across models in terms of semantic consistency and output reproducibility, even under identical conditions. While GPT-4.1 achieved both textual diversity and semantic consistency, the output repetition observed in Gemini-2.5-Flash and the semantic variability of Claude-Sonnet-4.6 each raise concerns about clinical reliability in distinct ways. Temperature settings were insufficient as a guarantee of consistency, and model selection was found to constitute a clinical rather than merely technical decision. The ceiling effect in safety expression revealed the limitations of current evaluation metrics and points to the need for more granular assessment frameworks. These findings support the importance of model-specific empirical validation and the development of standardized evaluation frameworks for the reliable clinical integration of LLM-based exercise prescription in sports and exercise medicine.

## Declaration of generative AI and AI-assisted technologies in the manuscript preparation process

During the preparation of this work, the authors used AI-based language tools to assist with translation and language refinement. After using these tools, the authors reviewed and edited the content as needed and take full responsibility for the content of the published article.

## Funding

This research received no external funding.

## Conflict of Interest

The author declares no conflict of interest.

## Reference

1. Akrimi, S., Schwensfeier, L., Düking, P., Kreutz, T., & Brinkmann, C. (2025). ChatGPT-4o-generated exercise plans for patients with type 2 diabetes mellitus: Assessment of their safety and other quality criteria by coaching experts. *Sports*, *13*(4), 92. https://doi.org/10.3390/sports13040092
2. American College of Sports Medicine. (2024). *ACSM's guidelines for exercise*

*testing and prescription* (12th ed.). Wolters Kluwer.

3. Atil, B., Aykent, S., Chittams, A., Fu, L., Passonneau, R. J., Radcliffe, E., Rajagopal, G. R., Sloan, A., Tudrej, T., Ture, F., Wu, Z., Xu, L., & Baldwin, B. (2025). Non-determinism of "deterministic" LLM system settings in hosted environments. In Proceedings of the 5th Workshop on Evaluation and Comparison of NLP Systems (pp. 135–148). Association for Computational Linguistics. https://arxiv.org/abs/2408.04667
4. Aydin, S., Karabacak, M., Vlachos, V., & Margetis, K. (2024). Large language models in patient education: A scoping review of applications in medicine. *Frontiers in Medicine*, *11*, 1477898. https://doi.org/10.3389/fmed.2024.1477898
5. Bishop, D. J., Beck, B., Biddle, S. J. H., Denay, K. L., Ferri, A., Gibala, M. J., Headley, S., Jones, A. M., Jung, M., Lee, M. J.-C., Moholdt, T., Newton, R. U., Nimphius, S., Pescatello, L. S., Saner, N. J., & Tzarimas, C. (2025). Physical activity and exercise intensity terminology: A joint ACSM expert statement and ESSA consensus statement. *Medicine & Science in Sports & Exercise*, *57*(11), 2599–2613. https://doi.org/10.1249/MSS.0000000000003795
6. Choi, M., Park, J., Lee, M., Beom, J., Jung, S. Y., & Lee, K. (2026). AI-generated exercise prescriptions for at-risk populations: Safety and feasibility of a large language model assessed by expert evaluation. *Journal of Clinical Medicine*, *15*(6), 2457. https://doi.org/10.3390/jcm15062457
7. Croxford, E., Gao, Y., First, E., Pellegrino, N., Schnier, M., Caskey, J., Oguss, M., Wills, G., Chen, G., Dligach, D., Churpek, M. M., Mayampurath, A., Liao, F., Goswami, C., Wong, K. K., Patterson, B. W., & Afshar, M. (2025). Evaluating clinical AI summaries with large language models as judges. *npj Digital Medicine*, *8*, 640. https://doi.org/10.1038/s41746-025-01648-5
8. Currier, B. S., D'Souza, A. C., Fiatarone Singh, M. A., Lowisz, C. V., Rawson, E. S., Schoenfeld, B. J., Smith-Ryan, A. E., Steen, J. P., Thomas, G. A., Triplett, N. T., Washington, T. A., Werner, T. J., & Phillips, S. M. (2026). American College of Sports Medicine position stand: Resistance training prescription for muscle function, hypertrophy, and physical performance in healthy adults. *Medicine & Science in Sports & Exercise*, *58*(4), 851–872. https://doi.org/10.1249/MSS.0000000000003897
9. Dergaa, I., Ben Saad, H., El Omri, A., Glenn, J. M., Clark, C. C. T., Washif, J. A., Guelmami, N., Hammouda, O., Al-Horani, R. A., Reynoso-Sánchez, L. F., Romdhani, M., Paineiras-Domingos, L. L., Vancini, R. L., Taheri, M., Mataruna-Dos-Santos, L. J., Trabelsi, K., Chtourou, H., Zghibi, M., Eken, Ö., Swed, S., Ben Aissa, M., Shawki, H. H., El-Seedi, H. R., Mujika, I., Seiler, S., Zmijewski, P., Pyne, D. B., Knechtle, B., Asif, I. M., Drezner, J. A., Sandbakk, Ø., & Chamari, K. (2024). Using artificial intelligence for exercise prescription in personalised health promotion: A critical evaluation of OpenAI's GPT-4 model. *Biology of Sport*, *41*(2), 221–241. https://doi.org/10.5114/biolsport.2024.133661
10. Düking, P., Sperlich, B., Voigt, L., Van Hooren, B., Zanini, M., & Zinner, C. (2024). ChatGPT generated training plans for runners are not rated optimal by coaching experts, but increase in quality with additional input information. *Journal of Sports Science and Medicine*, *23*, 56–65.
11. Enichen, E. J., Young, C. C., & Frates, E. P. (2025). The potential of AI to create personalized exercise plans. *Health Promotion Practice*. Advance online publication. https://doi.org/10.1177/15248399251394695
12. Festa, R. R., Jofré-Saldía, E., Candia, A. A., Monsalves-Álvarez, M., Flores-Opazo, M., Peñailillo, L., Marzuca-Nassr, G. N., Aguilar-Farias, N., Fritz-Silva, N., &

Cancino-Lopez, J. (2023). Next steps to advance general physical activity recommendations towards physical exercise prescription: A narrative review. *BMJ Open Sport & Exercise Medicine*, *9*, e001749. https://doi.org/10.1136/bmjsem-2023-001749

13. Garber, C. E., Blissmer, B., Deschenes, M. R., Franklin, B. A., Lamonte, M. J., Lee, I.-M., Nieman, D. C., & Swain, D. P. (2011). American College of Sports Medicine position stand: Quantity and quality of exercise for developing and maintaining cardiorespiratory, musculoskeletal, and neuromotor fitness in apparently healthy adults. *Medicine & Science in Sports & Exercise*, *43*(7), 1334–1359. https://doi.org/10.1249/MSS.0b013e318213fefb
14. He, Z., Wang, J., Zhang, B., & Li, Y. (2026). Knowledge-grounded large language model for personalized sports training plan generation. *Scientific Reports*, *16*, 6793. https://doi.org/10.1038/s41598-026-37075-z
15. Kim, B., Kang, J., Jung, Y. J., & Ahn, J. (2026). Generative and large-scale artificial intelligence in exercise and sports medicine: A narrative review. *The Asian Journal of Kinesiology*, *28*(1), 58–72. https://doi.org/10.15758/ajk.2026.28.1.58
16. Kim, J. H. (2026). Automated prescription of therapeutic exercise for shoulder impingement syndrome using literature-driven rule generation architecture. *Musculoskeletal Science and Practice*, *76*, 103520. https://doi.org/10.1016/j.msksp.2026.103520
17. Lai, X., Chen, J., Lai, Y., Huang, S., Cai, Y., Sun, Z., Wang, X., Pan, K., Gao, Q., & Huang, C. (2025a). Using large language models to enhance exercise recommendations and physical activity in clinical and healthy populations: Scoping review. *JMIR Medical Informatics*, *13*, e59309. https://doi.org/10.2196/59309
18. Lai, X., Lai, Y., Chen, J., Huang, S., Gao, Q., & Huang, C. (2025b). Evaluation strategies for large language model-based models in exercise and health coaching: Scoping review. *Journal of Medical Internet Research*, *27*, e79217. https://doi.org/10.2196/79217
19. Lai, X., Lai, Y., Chen, J., Huang, S., Gao, Q., & Huang, C. (2026). An AI-assisted adaptive boolean rubric for exercise prescription evaluation: A pilot validation study. *International Journal of Medical Informatics*, *207*, 106202. https://doi.org/10.1016/j.ijmedinf.2025.106202
20. Lee, K. (2026). Consistency of AI-generated exercise prescriptions: A repeated generation study using a large language model. *arXiv preprint arXiv:2604.11287*. https://arxiv.org/abs/2604.11287
21. Li, G., Li, H., Su, Y., Li, Y., Jiang, S., & Zhang, G. (2025). GPT-4 as a virtual fitness coach: A case study assessing its effectiveness in providing weight loss and fitness guidance. *BMC Public Health*, *25*, 2466. https://doi.org/10.1186/s12889-025-23666-6
22. Li, H., Dong, Q., Chen, J., Su, H., Zhou, Y., Ai, Q., Ye, Z., & Liu, Y. (2024). LLMs-as-judges: A comprehensive survey on LLM-based evaluation methods. *arXiv preprint arXiv:2412.05579*. https://arxiv.org/abs/2412.05579
23. Meng, X., Yan, X., Zhang, K., Liu, D., Cui, X., Yang, Y., Zhang, M., Cao, C., Wang, J., Wang, X., Gao, J., Wang, Y.-G.-S., Ji, J.-M., Qiu, Z., Li, M., Qian, C., Guo, T., Ma, S., Wang, Z., . . . Tang, Y.-D. (2024). The application of large language models in medicine: A scoping review. *iScience*, *27*, 109713. https://doi.org/10.1016/j.isci.2024.109713
24. Nduka, T. C., Ndakotsu, A., Nriagu, V. C., Karikalan, S., Abdulkareem, L., Omede, F. O., & Bob-Manuel, T. (2025). AI-generated diet and exercise recommendations for cardiovascular health compared to established cardiology society guidelines. *Cureus*, *17*(8), e90968. https://doi.org/10.7759/cureus.90968
25. Negra, Y., Sammoud, S., Bouguezzi, R., Markov, A., Capranica, L., Müller, P., & Chaabene, H. (2026). Effects of a ChatGPT-generated eccentric training programme on speed, change of direction, agility, and jumping performance in U14 tennis players: A non-randomised controlled study. Journal of Sports Sciences. Advance online publication. https://doi.org/10.1080/02640414.2026.2653295
26. Philuek, P., Kusump, S., Sathianpoonsook, T., Jansupom, C., Sawanyawisuth, P., Sawanyawisuth, K., & Chainarong, A. (2025). The effects of chat GPT generated exercise program in healthy overweight young adults: A pilot study. *Journal of Human Sport and Exercise*, *20*, 169–179. https://doi.org/10.14198/jhse.2025.201.15
27. Puce, L., Bragazzi, N. L., Currà, A., & Trompetto, C. (2025). Harnessing generative artificial intelligence for exercise and training prescription: Applications and implications in sports and physical activity—A systematic literature review. *Applied Sciences*, *15*(7), 3497. https://doi.org/10.3390/app15073497
28. Raza, M. M., Venkatesh, K. P., & Kvedar, J. C. (2024). Generative AI and large language models in health care: Pathways to implementation. *npj Digital Medicine*, *7*, 62. https://doi.org/10.1038/s41746-023-00988-4
29. Reimers, N., & Gurevych, I. (2019). Sentence-BERT: Sentence embeddings using Siamese BERT-networks. In Proceedings of the 2019 Conference on Empirical Methods in Natural Language Processing and the 9th International Joint Conference on Natural Language Processing (EMNLP-IJCNLP) (pp. 3982–3992). https://doi.org/10.18653/v1/D19-1410
30. Schoenfeld, B. J., Grgic, J., Van Every, D. W., & Plotkin, D. L. (2021). Loading recommendations for muscle strength, hypertrophy, and local endurance: A re-examination of the repetition continuum. *Sports*, *9*(2), 32. https://doi.org/10.3390/sports9020032
31. Schütze, K., Shehatha, R., Beer, K., Needham, M., Smith, T., Bagg, M., Doverty, A., & Cooper, I. (2026). Evaluating ChatGPT’s advice and recommendations regarding exercise for people with inclusion body myositis. *Neuromuscular Disorders*, *62*, 106418. https://doi.org/10.1016/j.nmd.2026.106418
32. Shin, D., Hsieh, G., & Kim, Y. H. (2025). PlanFitting: Personalized exercise planning with large language model-driven conversational agent. In Proceedings of the 7th ACM Conference on Conversational User Interfaces (CUI ’25). https://doi.org/10.1145/3719160.3736607
33. Shyr, C., Ren, B., Hsu, C.-Y., Yan, C., Tinker, R. J., Cassini, T. A., Hamid, R., Wright, A., Bastarache, L., Peterson, J. F., Malin, B. A., & Xu, H. (2025). A statistical framework for evaluating the repeatability and reproducibility of large language models. *medRxiv*. https://doi.org/10.1101/2025.08.06.25333170
34. Song, Y., Wang, G., Li, S., & Lin, B. Y. (2024). The good, the bad, and the greedy: Evaluation of LLMs should not ignore non-determinism. *arXiv preprint arXiv:2407.10457*. https://arxiv.org/abs/2407.10457
35. Washif, J., Pagaduan, J., James, C., Dergaa, I., & Beaven, C. (2024). Artificial intelligence in sport: Exploring the potential of using ChatGPT in resistance training prescription. *Biology of Sport*, *41*(2), 209–220. https://doi.org/10.5114/biolsport.2024.132987
36. Wataoka, K., Takahashi, T., & Ri, R. (2024). Self-preference bias in LLM-as-a-judge. *arXiv preprint arXiv:2410.21819*. https://arxiv.org/abs/2410.21819
37. Zaleski, A. L., Berkowsky, R., Craig, K. J. T., & Pescatello, L. S. (2024).

Comprehensiveness, accuracy, and readability of exercise recommendations provided by an AI-based chatbot: Mixed methods study. *JMIR Medical Education*, *10*, e51308. https://doi.org/10.2196/51308

38. Zhang, Y.-F., & Liu, X.-Q. (2024). Using ChatGPT to promote college students' participation in physical activities and its effect on mental health. *World Journal of Psychiatry*, *14*, 330–341. https://doi.org/10.5498/wjp.v14.i2.330
39. Zheng, L., Chiang, W.-L., Sheng, Y., Zhuang, S., Wu, Z., Zhuang, Y., Lin, Z., Li, Z., Li, D., Xing, E. P., Zhang, H., Gonzalez, J. E., & Stoica, I. (2023). Judging LLM-as-a-judge with MT-bench and chatbot arena. *Advances in Neural Information Processing Systems*, *36*. https://arxiv.org/abs/2306.05685

# Supplementary Material 1. Exercise Prescription Generation Prompt

## 1. Prompt Template

The following prompt template was used to generate exercise prescriptions for all six clinical scenarios. The placeholder [CLINICAL CASE] was replaced with the corresponding clinical case description for each scenario.

```
INSTRUCTION_TEMPLATE =
"""Based on [CLINICAL CASE], please develop a 12-week exercise
program.
Ensure that the plan adheres to international exercise
guidelines
(e.g., ACSM Guidelines for Exercise Testing and Prescription;
ADA Standards of Care; OARSI Guidelines), and clearly indicate
contraindications and precautions. Include aerobic,
resistance,
balance, and flexibility exercises, and specify exercise
intensity
using RPE or heart rate-based methods for aerobic exercise,
and %1RM for resistance exercise."""
```

Note. This prompt was adapted from the guideline-based instruction (JCM Prompt 2) used in a previous study (Choi et al., 2026) with the following modifications: (1) resistance exercise intensity notation (%1RM) was added; (2) the 'ACSM Guidelines for Cancer Survivors' was removed as it was not applicable to all scenarios; and (3) the 'if applicable' condition was removed to ensure that all four exercise components (aerobic, resistance, balance, and flexibility) were consistently included across all outputs.

## 2. Clinical Case Descriptions Used as Input

The following clinical case descriptions were substituted into the [CLINICAL CASE] placeholder of the prompt template above. All cases are hypothetical and were constructed without the use of real patient information.

### Case 1. Type 2 Diabetes Mellitus + Obesity

Participant Profile

Male, 55 years old, 7-year history of type 2 diabetes mellitus (on metformin + sulfonylurea), BMI 31.2. Limited exercise experience. Comorbidities: mild peripheral neuropathy, no diabetic retinopathy. Baseline physical activity level: low. Goals: weight reduction and improved glycemic control.

### Case 2. Knee Osteoarthritis + Fall Risk

Participant Profile

Female, 70 years old, 5-year diagnosis of knee osteoarthritis, BMI 28. Able to walk but experiences knee pain; one prior fall incident. No other comorbidities. No medications reported. Baseline physical activity level: low. Goal: pain reduction, maintenance of walking ability, and fall prevention.

**Case 3. Post-Colon Cancer Surgery Recovery**

Participant Profile

Male, 60 years old, 6 months post-colon cancer surgery; chemotherapy completed 2 months prior. Deconditioned; currently able to walk for approximately 15 minutes. No other comorbidities. Baseline physical activity level: low. Goal: physical recovery, fatigue reduction, and improvement of lifestyle habits.

**Case 4. Hypertension + Type 2 Diabetes + Obesity (Multimorbidity)**

Participant Profile

Female, 68 years old, diagnosed with hypertension, type 2 diabetes mellitus, and obesity. On antihypertensive and antidiabetic medication. Comorbidities: dizziness during daily activities, mild exertional dyspnea. Baseline physical activity level: low. Goal: safe improvement of cardiovascular fitness, glycemic control, and gradual weight reduction.

**Case 5. Healthy Adult – Fat Loss and Cardiovascular Endurance**

Participant Profile

Female, 28 years old, no clinical conditions. Moderate to high physical activity level (regular aerobic activity 3–4 times per week). No functional limitations or contraindications. No medications. Goal: reduction of body fat and improvement of cardiovascular endurance.

**Case 6. Healthy Adult – Muscle Hypertrophy and Strength**

Participant Profile

Male, 30 years old, no clinical conditions. High physical activity level (regular resistance training ≥4 times per week). Prior resistance training experience; no functional limitations or contraindications. No medications. Goal: increase muscle mass (hypertrophy) and improve overall strength.

## Supplementary Material 2. Evaluation Prompts and Classification Criteria

### 1. FITT Structural Classification

#### 1.1 Prompt Used for FITT Classification (Claude Sonnet 4.6)

The following prompt was submitted to Claude Sonnet 4.6 (Anthropic) for each of the 120 preprocessed outputs. The placeholder [EXERCISE PRESCRIPTION TEXT] was replaced with the corresponding output text.

```
FITT_PROMPT =
"""Classify the FITT components based on the initial prescription
(Weeks 1–4) from the following exercise prescription.

Classification criteria:
- Frequency: Weekly exercise frequency (e.g., 3 days/week)
- Intensity_Aerobic: low / moderate / high / unclassifiable
  * low:      RPE <12, %HRmax <64%, HRR <40%, MET <3
  * moderate: RPE 12-13, %HRmax 64-76%, HRR 40-59%, MET 3-5.9
  * high:     RPE ≥14, %HRmax ≥77%, HRR ≥60%, MET ≥6
- Intensity_Resistance: low / moderate / high / unclassifiable
  * low:      <50% 1RM
  * moderate: 50-69% 1RM
  * high:     ≥70% 1RM
- Time: Duration per session (minutes)
- Type: Exercise modality (e.g., walking, cycling, resistance
training)

Output ONLY in the following JSON format. Do not add any other text:
{
  "Frequency": "",
  "Intensity_Aerobic": "",
  "Intensity_Resistance": "",
  "Time": "",
  "Type": ""
}

Exercise prescription:
[EXERCISE PRESCRIPTION TEXT]"""
```

#### 1.2 Intensity Classification Criteria

The intensity classification criteria applied in this study are summarized in the table below. Aerobic intensity was classified with reference to the ACSM guidelines (Garber et al., 2011) and the joint ACSM–ESSA expert consensus statement (Bishop et al., 2025). Resistance exercise intensity was classified based on the repetition continuum framework (Schoenfeld et al., 2021) and the ACSM position stand on resistance training prescription (Currier et al., 2026). When the relevant metric was not explicitly stated, descriptive expressions were estimated and classified according to ACSM criteria and labeled as 'estimated.' Cases that could not be classified were recorded as 'unclassifiable.'

| Level | Aerobic Exercise | Resistance Exercise |
| --- | --- | --- |
| Low | RPE <12 / %HRmax <64% / HRR <40% / <3 METs | <50% 1RM |
| Moderate | RPE 12–13 / %HRmax 64–76% / HRR 40–59% / 3–5.9 METs | 50–69% 1RM |
| High | RPE ≥14 / %HRmax ≥77% / HRR ≥60% / ≥6 METs | ≥70% 1RM |
| Unclassifiable | Expression present but cannot be mapped to above criteria | Expression present but %1RM not stated or estimable |

Note. RPE = rating of perceived exertion (Borg 6–20 scale); HRmax = maximum heart rate; HRR = heart rate reserve; MET = metabolic equivalent of task; 1RM = one-repetition maximum.

## 2. Safety Expression Consistency Evaluation

### 2.1 Prompt Used for Safety Evaluation (Claude Sonnet 4.6)

The following prompt was submitted to Claude Sonnet 4.6 (Anthropic) for binary inclusion assessment of safety-related expressions in each output.

```
SAFETY_PROMPT =
"""Evaluate the presence or absence of safety-related expressions
in the following exercise prescription.

Evaluation criteria (1 = included, 0 = not included):
- Contraindication: Whether exercise contraindications are explicitly
stated
- Precaution:       Whether precautionary statements are included
- Monitoring:       Whether symptom monitoring instructions are
provided
- Warning:          Whether explicit risk warnings are included

Output ONLY in the following JSON format. Do not add any other text:
{
  "Contraindication": 0,
  "Precaution": 0,
  "Monitoring": 0,
```

```
  "Warning": 0
}

Exercise prescription:
[EXERCISE PRESCRIPTION TEXT]"""
```

### 2.2 Safety Category Definitions and Classification Criteria

The four safety evaluation categories and their operational definitions are presented below. Contraindications and precautions were based on concepts used in the safety evaluation rubric of the previous study (Choi et al., 2026). Symptom monitoring adopted terminology from the ACSM Guidelines for Exercise Testing and Prescription (ACSM, 2024). Risk warnings were operationally defined for the purposes of this study with reference to clinical safety concepts in the same guidelines.

| Category | Definition | Scoring |
| --- | --- | --- |
| Contraindication | Explicit mention of activities or conditions that are absolutely or relatively contraindicated for the given clinical case | 1 = included, 0 = not included |
| Precaution | Statements advising caution, modification, or gradual progression due to clinical risk factors | 1 = included, 0 = not included |
| Monitoring | Instructions for symptom monitoring during or after exercise (e.g., heart rate, blood glucose, pain level, dyspnea) | 1 = included, 0 = not included |
| Warning | Explicit risk warnings advising when to stop exercise or seek medical attention | 1 = included, 0 = not included |

Note. All four categories were evaluated as binary items (1 = included, 0 = not included). In addition to binary scoring, sentence-level counts per category were also computed using Claude Sonnet 4.6 to quantify the extent of safety expression across outputs.

## 3. Preprocessing Prompt

### 3.1 Prompt Used for Output Preprocessing (Claude Sonnet 4.6)

Prior to SBERT-based semantic similarity analysis, a standardized preprocessing prompt was applied to all 120 raw outputs using Claude Sonnet 4.6 to extract only the exercise prescription body, excluding formatting elements such as greetings, closing remarks, tables, and bullet point headers unrelated to the prescription content.

```
PREPROCESSING_PROMPT =
"""From the following text, extract only the exercise prescription
body.
```

Remove all greetings, closing remarks, and formatting elements (e.g., table headers, bullet symbols) that are unrelated to the prescription content itself. Preserve all specific exercise recommendations, intensity values, frequency, duration, and safety-related content exactly as written.

Output only the extracted prescription text. Do not add any explanations or additional content.

Text:
[RAW OUTPUT TEXT]"""